\documentclass{article} 
\usepackage{iclr2027_conference,times}

\usepackage{amsmath,amsfonts,bm}

\def\eqref#1{equation~\ref{#1}}

\def\1{\bm{1}}

\DeclareMathAlphabet{\mathsfit}{\encodingdefault}{\sfdefault}{m}{sl}
\SetMathAlphabet{\mathsfit}{bold}{\encodingdefault}{\sfdefault}{bx}{n}

\usepackage[dvipsnames]{xcolor}
\usepackage{hyperref}
\hypersetup{colorlinks=true, linkcolor=BrickRed, citecolor=NavyBlue, urlcolor=Blue}
\usepackage{url}
\usepackage{graphicx}
\usepackage{placeins}
\usepackage{soul}

\title{Representational Simplicity and Circuit Size Dissociate in a Threshold-Dependent Way: A Controlled Test via Adversarial Training}

\author{Adam Elimadi \\
Independent \\
Cambridge, MA \\
\texttt{elimadadam@gmail.com}
}

\iclrfinalcopy
\makeatletter
\def\@maketitle{\vbox{\hsize\textwidth
  {\centering\LARGE\sc \@title\par}
  \lhead{Preprint}
  {\centering\begin{tabular}[t]{c}\bf\rule{\z@}{24pt}\@author\end{tabular}\par}
  \vskip 0.3in minus 0.1in}}
\makeatother

\begin{document}
\lhead{Preprint}

\maketitle
\lhead{Preprint}\thispagestyle{fancy}

\begin{abstract}
Sparse-autoencoder decomposability and concentrated feature attribution are increasingly treated as evidence that a model's computation is easier to reverse-engineer. Whether this representational and attributional cleanliness actually predicts a smaller or more tractable causal circuit remains an open question.
We test this directly using adversarial training as a controlled instrument: it reliably reshapes internal representations, but this alone does not constitute a test of circuit size.
We investigate this question through \emph{reverse-engineering complexity}: the causal structure required to recover a model's behavior at a fixed level of faithfulness. To our knowledge, this is the first controlled empirical test of whether representational or attributional simplicity translates into causal simplicity at the circuit level.
Starting from the same pretrained GPT-2 Small checkpoint, we apply matched standard and adversarial continual training, requiring both conditions to retain competence on indirect object identification and pass independent robustness verification before comparing mechanisms.
We then compare the models along three complementary axes: sparse-autoencoder decomposability, SAE feature engagement in task attribution, and the size of faithful circuits recovered from the raw computational graph.
The robust model is more SAE-decomposable and engages fewer SAE features in task attribution. Circuit size is regime-dependent: on competence-matched IOI, standard leads or ties below 85\% faithfulness, but robust needs substantially fewer edges at high faithfulness (90\%, 95\%). The high-faithfulness advantage holds for every qualified robust model in a seven-point adversarial-strength sweep and for a second corpus, as do the representational trends.
\end{abstract}

\section{Introduction}
\label{sec:introduction}
 
Mechanistic interpretability seeks to reverse-engineer the computations implemented by neural networks.
Prior work has identified circuits for specific behaviors, including indirect object identification (IOI) and Greater-Than in GPT-2 Small
\citep{wang2022interpretability,hanna2023greaterthan}.
Yet models that exhibit the same behavior need not implement it through mechanisms of comparable complexity.
Their solutions may differ in how distributed, redundant, or causally entangled they are, and therefore in how difficult they are to recover.

\begin{figure}[t]
\centering
\includegraphics[width=\textwidth]{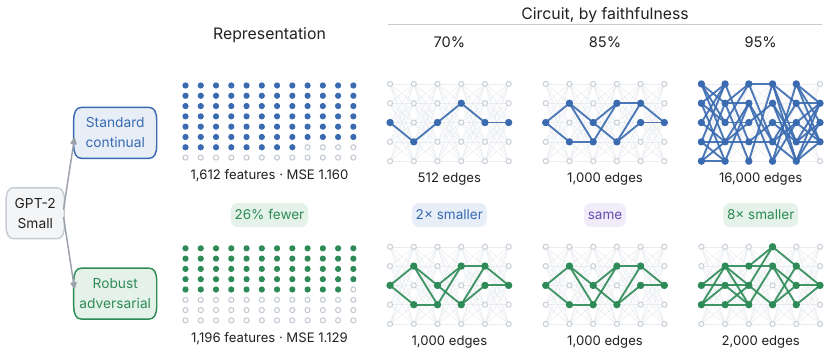}
\caption{Simpler representations do not imply uniformly smaller circuits. Starting from one GPT-2 Small checkpoint, we train standard and adversarially robust models with matched IOI competence. The robust model engages 26\% fewer SAE features and is reconstructed more accurately, yet its circuit is not uniformly smaller: at 70\% faithfulness the standard model's partial subgraph is the smaller one, at 85\% both need the same 1{,}000 edges, and the robust circuit becomes markedly smaller only at 90\% and 95\%. Badges give the size gap, colored by the model that needs fewer features or edges; circuit drawings are schematic, with counts as measured. This pattern replicates across the adversarial-strength sweep and a second corpus.}
\label{fig:overview}
\end{figure}

Adversarial training provides a natural setting in which to study this variation.
It can substantially alter the features and representations learned by a model
\citep{tsipras2019robustness,engstrom2019adversarialrobustnesspriorlearned},
and has been associated with more interpretable gradients and saliency maps
\citep{kim2019bridgingadversarialrobustnessgradient,etmann2019connectionadversarialrobustnesssaliency}.
Recent work also links adversarial vulnerability to superposition and interference between overlapping features
\citep{gorton2025adversarialexamplesbugssuperposition,bereska2025superpositionlossycompressionmeasure,stevinson2026adversarialattacksleverageinterference},
and the reduction in superposition under adversarial training has been traced to robust models representing fewer features, with the dropped features being predominantly non-robust \citep{elimadi2026robustness}.
If robust training reduces this entanglement, it may make learned computations easier to isolate.
Conversely, it may induce representational drift, redundancy, or compensatory pathways that make circuits harder to recover.
Moreover, representational simplicity need not imply causal simplicity
\citep{xu2026pattern}.
Adversarial training lets us test this directly: it reliably alters representations while task behavior can be held fixed.
 
We measure \emph{reverse-engineering complexity} as the size of the smallest faithful circuit that recovers a retained behavior, not as easier SAE reconstruction or more concentrated attribution. Our design (Section~\ref{sec:experimental-design}) combines matched continual training from a shared checkpoint, competence and robustness qualification, and three axes of simplicity; it separates changes in representation geometry from changes in causal organization and tests whether apparent simplicity is instead explained by redundant or compensatory computation.
 
\paragraph{Contributions.}
\begin{itemize}
    \item \textbf{Representational simplicity does not uniformly predict causal simplicity}: adversarial training increases SAE decomposability and reduces the number of active attribution features, yet the faithful circuit recovered by edge attribution patching is no smaller at low-to-moderate faithfulness thresholds; whether the proxy holds depends on the faithfulness required.

    \item \textbf{Robust models have substantially smaller high-faithfulness circuits}: to our knowledge, this is the first controlled evidence that adversarial training, which has no interpretability objective and can expand a model's feature set \citep{bereska2025superpositionlossycompressionmeasure}, yields substantially smaller circuits for a behavior (IOI) when high faithfulness is required: $4\times$ fewer edges at 90\% and $8\times$ fewer at 95\% for the primary pair, replicating ($2$--$8\times$) across all qualified robust models in a seven-model adversarial-strength sweep and a second corpus.

    \item We corroborate this with $\alpha$-ReQ, the decay rate of the residual-stream covariance spectrum, an SAE-free measure of representational simplicity: it rises with adversarial strength across a seven-point sweep, confirming that the representational trend is not an artifact of the SAE training procedure.
\end{itemize}

\section{Experimental Design}
\label{sec:experimental-design}

\subsection{Overview}

We test whether adversarial training changes how readily a learned computation can be reverse engineered while controlling for initialization, data, training budget, and task performance. We continually train a standard and a family of adversarially trained GPT-2 Small models from the same pretrained checkpoint and on the same token stream. We first qualify candidate pairs by checking task retention and robustness under a multi-step attack. We then compare the selected standard and robust models along three progressively stronger axes: (i) how accurately matched sparse autoencoders (SAEs) reconstruct their internal activations, (ii) how many SAE features engage in task attribution, and (iii) how large a subgraph must be to reach a given faithfulness level. The first two axes measure decomposability and attribution structure; only the third directly measures reverse-engineering complexity.

Our primary matched-competence task is indirect object identification (IOI), for which all trained models retain near-ceiling performance. We additionally evaluate the Greater-Than (GT) task, which is not competence-matched between models.

\subsection{Matched continual training}
\label{sec:continual-training}

All models initialize from the official 124M-parameter GPT-2 Small checkpoint. Each model receives the same 1B-token stream in the same order and updates all parameters for 1,908 optimization steps. The primary comparison uses OpenWebText; Section~\ref{sec:generalization} reports an identical standard-robust pair trained on FineWeb instead, to test whether the corpus drives the result. The standard arm minimizes next-token cross-entropy. The adversarial arms optimize
\begin{equation}
    \mathcal{L}
    = (1-\alpha)\mathcal{L}_{\mathrm{clean}}
    + \alpha\mathcal{L}_{\mathrm{adv}},
    \label{eq:training-objective}
\end{equation}
where $\mathcal{L}_{\mathrm{adv}}$ is evaluated after a ten-step projected-gradient attack on token-embedding activations. Perturbations are applied after the token lookup and before positional embeddings. For every non-special token, the attack is constrained to an $\ell_2$ ball with radius $\epsilon_{\mathrm{rel}}$ times the mean non-special-token embedding norm in the micro-batch. We use random initialization, per-token normalized gradient ascent, and projection after every step; the special token is never perturbed. This continuous embedding-space construction follows the general setup of \citet{xhonneux2024efficient}.

We sweep $\epsilon_{\mathrm{rel}}\in\{0.05,0.075,0.10\}$ and $\alpha\in\{0.2,0.5\}$, producing six adversarial candidates and one standard control. All runs use AdamW, a peak learning rate of $5\times10^{-5}$ with 100 warmup steps and cosine decay, a global batch of 524,288 tokens, gradient clipping at 1.0, and bfloat16 computation. Appendix~\ref{app:training-details} gives the complete optimization and attack specification.

\subsection{Tasks and qualification criteria}
\label{sec:qualification-method}

For IOI, we evaluate 1,000 fixed prompts balanced between ABBA and BABA templates \citep{wang2022interpretability}. The behavioral score is the logit difference between the correct and incorrect names at the final position, and accuracy is the fraction of prompts with positive logit difference. The standard model must achieve at least 90\% accuracy and each adversarial candidate at least 85\%, floors that allow for ordinary continual-training drift while remaining far above the level indicating circuit absence (55.4\% for a from-scratch control that failed to develop the behavior, Appendix~\ref{app:training-details}).

For robustness, we evaluate held-out OpenWebText under both a single-step attack and the ten-step attack used during training. The masking sanity check requires
\begin{equation}
    \left|\mathcal{L}_{\mathrm{PGD10}}-\mathcal{L}_{\mathrm{single}}\right|
    \leq 0.10 \text{ nats},
    \label{eq:masking-gate}
\end{equation}
which guards against selecting a model that appears robust only under a weak attack \citep{athalye2018obfuscated}. Among candidates passing this check and the IOI floor, we rank robustness by attack susceptibility,
\begin{equation}
    \Delta_{\mathrm{adv}}
    = \mathcal{L}_{\mathrm{PGD10}}-\mathcal{L}_{\mathrm{clean}},
    \label{eq:attack-susceptibility}
\end{equation}
where lower values indicate a smaller attack-induced loss increase. We select the qualified candidate with the lowest $\Delta_{\mathrm{adv}}$. The single-step versus PGD-10 gap is a masking diagnostic, not a robustness score; the non-robust standard control is therefore not required to pass it.

We also evaluate 1,000 GT prompts generated with the official YearDataset \citep{hanna2023greaterthan}. For a two-digit threshold $YY$, the probability-difference score is
\begin{equation}
    \operatorname{PD}
    = \sum_{y>YY}p(y)-\sum_{y\leq YY}p(y),
    \label{eq:gt-pd}
\end{equation}
where the sums use the full-softmax probabilities of the 100 two-digit year tokens. We report paired differences from the standard model on identical prompts. GT is not a qualification gate because no established performance floor analogous to the IOI accuracy threshold exists for this task.

\subsection{Axis 1: SAE decomposability}
\label{sec:sae-method}

We train matched BatchTopK SAEs \citep{bussmann2024batchtopk} on the residual stream entering transformer block 8, following standard practice of choosing a layer near the end of the network that contains diverse features without being specialized for final-layer output computation \citep{gao2025scaling}. Each SAE has 12,288 features, a 16-fold expansion over the 768-dimensional residual stream, and an average $L_0$ of 32 selected features per token. Training uses 300M post-training OpenWebText tokens disjoint from model training; evaluation uses a subsequent 10M-token partition. The standard and robust SAEs share the same architecture, sparsity target, data volume, optimizer, number of updates, and evaluation protocol. We train three seeds per condition for the principal reconstruction comparison.

The primary endpoint is mean squared reconstruction error in the original residual space on the held-out evaluation partition. We also report normalized-space training loss, fixed-threshold inference error, dead-feature counts, and the between-seed dispersion. The robust SAE's lower error at fixed dictionary size and sparsity indicates that its activation distribution is more compressible by this SAE family.

\subsection{Axis 2: SAE feature engagement and selectivity}
\label{sec:attribution-method}

For each SAE (seed 42 primary; seeds 43 and 44 replicated), we attribute the IOI and GT scores to its features using node-level attribution patching \citep{syed2024attribution,marks2025sparsefeaturecircuits}. We splice the reconstruction $\hat{x}$ into the residual stream as
\begin{equation}
    \tilde{x}=x+\hat{x}-\operatorname{sg}(\hat{x}),
    \label{eq:straight-through-splice}
\end{equation}
such that the forward pass is exactly the unmodified model while gradients flow through the SAE features. For feature $i$, target attribution is
\begin{equation}
    A_i
    = \mathbb{E}_{x}\left[
      \sum_{t\in\mathcal{T}}
      \frac{\partial Y}{\partial f_{i,t}}
      \left(f_{i,t}^{\mathrm{clean}}-f_{i,t}^{\mathrm{corrupt}}\right)
    \right],
    \label{eq:feature-attribution}
\end{equation}
where $Y$ is IOI logit difference or GT probability difference. We define the intervention position set $\mathcal{T}$ per task: $\mathcal{T}_{\mathrm{IOI}} = \{S1, IO, S2, \mathrm{END}\}$, the three name positions and the final position; $\mathcal{T}_{\mathrm{GT}} = \{\text{start-year span}, \mathrm{END}\}$, the token span stating the prompt's start year and the final position. Corrupted prompts swap the two IOI names or use the official GT counterfactual construction.

Our primary cross-model statistic is candidate pool size: the number of features active at task-relevant positions under an identical existence rule and identical prompts, measured before any ranking or attribution computation. As a secondary statistic, after sorting existing features by $|A_i|$, we report the smallest number required to account for 50\%, 80\%, and 90\% of total absolute attribution. As a complementary check on whether a feature's causal role is specific to the task or also shapes the model's broader output, we zero each pool feature on the same prompts and measure the shift in the full next-token distribution rather than the task readout alone (mean ablation-KL, Appendix~\ref{app:selectivity-controls}).

\subsection{Axis 3: faithful circuit recovery}
\label{sec:circuit-method}

We recover task circuits with edge attribution patching (EAP-IG); a pilot of EAP-GP, a more recent variant designed to reduce gradient saturation, is reported in Appendix~\ref{app:circuit-protocol} \citep{syed2024attribution,zhang2025eapgp}. For each task and model, we define edge-native faithfulness at $k$ retained edges as
\begin{equation}
    F(k) = \frac{Y(\mathrm{circuit}_k) - Y(\mathrm{corrupt})}{Y(\mathrm{clean}) - Y(\mathrm{corrupt})},
    \label{eq:faithfulness}
\end{equation}
where $\mathrm{circuit}_k$ patches in the top-$k$ edges by $|\mathrm{score}|$ at their clean values and corrupt-patches every other edge to its value under the task's counterfactual distribution (name-swap for IOI, the official counterfactual construction for GT, Section~\ref{sec:qualification-method}); $Y(\mathrm{clean})$ and $Y(\mathrm{corrupt})$ are the task metric (IOI logit difference or GT probability difference) on the fully clean and fully corrupted graph, so $F=0$ reproduces the fully corrupted baseline and $F=1$ fully recovers clean performance. We distinguish two estimands on the same $F(k)$ curve. Partial-subgraph recoverability is the area under $F(k)$ across the full edge grid, including $k$ where $F<85\%$. Faithful-circuit size is the first grid $k$ at which $F(k)$ reaches a target in the circuit regime $F\geq85\%$; we report that crossing at several targets, not at a single canonical cutoff. The causal claim in this paper is about this estimand. AUC is reported as a descriptor of partial recovery, not as a circuit-size statistic. Conditional co-ablation tests whether apparently weak components become necessary after substitutes are removed, guarding against redundancy and self-repair \citep{gong2026coax,hanna2024have,rushing2024selfrepair}.

\section{Results}
\label{sec:results}

\FloatBarrier

\subsection{Adversarial continual training yields a qualified robust comparison}
\label{sec:qualification-results}

All seven continually trained models retain IOI performance: accuracy ranges from 99.4\% to 99.6\%, compared with 99.7\% for the original checkpoint. Four adversarial candidates also pass the masking sanity check. Among them, \texttt{ft-e100-a05} has the lowest attack susceptibility, reducing $\Delta_{\mathrm{adv}}$ from 1.642 nats in the standard control to 0.362 while retaining 99.4\% IOI accuracy. We therefore use \texttt{ft-clean} and \texttt{ft-e100-a05} as the principal standard-robust pair.

\label{sec:behavioral-cost}
The near-constant IOI accuracy does not extend to GT. GT probability difference decreases smoothly as attack susceptibility falls across the sweep (Table~\ref{tab:qualification}). The selected robust model scores 0.442, compared with 0.671 for the standard model, with a paired prompt-level difference of $-0.229\pm0.010$. By construction, PD lies in $[-1,1]$, with values near zero indicating no informative signal; both models remain well above that floor, confirming the behavior persists substantially intact despite the competence gap between models on this task.

\begin{table}[!ht]
\centering
\caption{Qualification sweep and secondary GT performance. The gap is $|\mathcal{L}_{\mathrm{PGD10}}-\mathcal{L}_{\mathrm{single}}|$; $\Delta_{\mathrm{adv}}$ is the attack-induced loss increase, so lower is more robust. Asterisks mark adversarial candidates that pass both qualification gates.}
\label{tab:qualification}
\scriptsize
\setlength{\tabcolsep}{3.2pt}
\begin{tabular}{lrrrrrr}
\hline
Run & $\epsilon_{\mathrm{rel}}$ & $\alpha$ & IOI & Gap & $\Delta_{\mathrm{adv}}$ & GT PD \\
\hline
\texttt{ft-clean}     & 0     & 0   & .996 & .907 & 1.642 & .671 \\
\texttt{e050-a02}     & .050  & .2  & .994 & .153 & .655  & .609 \\
\texttt{e050-a05}$^*$ & .050  & .5  & .995 & .087 & .494  & .559 \\
\texttt{e075-a02}     & .075  & .2  & .994 & .104 & .556  & .584 \\
\texttt{e075-a05}$^*$ & .075  & .5  & .994 & .057 & .413  & .500 \\
\texttt{e100-a02}$^*$ & .100  & .2  & .995 & .079 & .495  & .554 \\
\textbf{\texttt{e100-a05}$^*$} & \textbf{.100} & \textbf{.5} & \textbf{.994} & \textbf{.041} & \textbf{.362} & \textbf{.442} \\
\hline
\end{tabular}
\end{table}

\FloatBarrier

\subsection{Robust activations are more sparsely reconstructable}
\label{sec:sae-results}

At equal dictionary size and sparsity, the SAE trained on robust-model activations reconstructs the held-out residual stream more accurately. Across three SAE seeds, original-space MSE is $1.1293\pm0.0009$ for the robust model and $1.1601\pm0.0022$ for the standard model, a 2.65\% reduction with non-overlapping seed ranges. On position-matched evaluation blocks, the standard-minus-robust difference is $0.0323\pm0.0007$. The same ordering holds under fixed-threshold inference and in normalized activation space (Table~\ref{tab:sae-reconstruction}), and at most one of the three seeds per condition has a single dead feature (Appendix~\ref{app:sae-details}).

\begin{table}[!ht]
\centering
\caption{Matched SAE reconstruction at the residual stream entering block 8. Lower is better. The original-space entry reports mean $\pm$ standard deviation across three SAE seeds; the remaining entries use seed 42.}
\label{tab:sae-reconstruction}
\small
\begin{tabular}{lrr}
\hline
Metric & Standard & Robust \\
\hline
Original-space MSE & $1.1601\pm0.0022$ & $\mathbf{1.1293\pm0.0009}$ \\
Fixed-threshold MSE & 1.1546 & \textbf{1.1262} \\
Normalized-space loss & 0.0877 & \textbf{0.0768} \\
Dead features & 0 & 0 \\
\hline
\end{tabular}
\end{table}

\FloatBarrier

\subsection{Robust task attribution engages fewer SAE features}
\label{sec:attribution-results}

Two effects distinguish the models, both consistent across all three SAE seeds and both tasks. First, under an identical existence rule and identical prompts, fewer robust-SAE features are active at task-relevant positions: 1,196 versus 1,612 candidates for IOI (a 26\% reduction) and 360 versus 563 for GT (36\%; Table~\ref{tab:attribution-concentration}). This measurement precedes any ranking or attribution computation.

\begin{figure}[!ht]
\centering
\includegraphics[width=\textwidth]{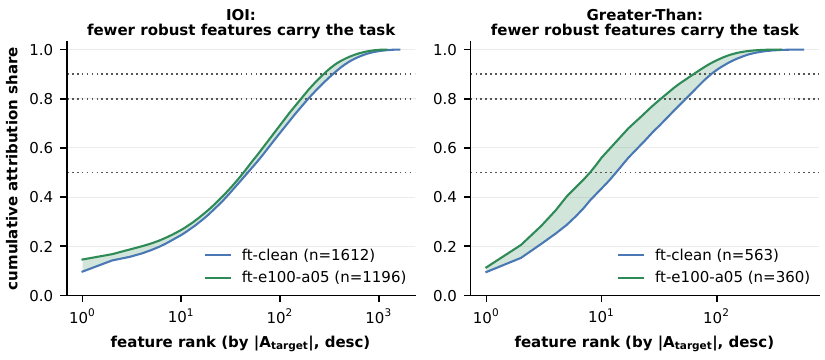}
\caption{Cumulative attribution share by feature rank (descending $|A_i|$), IOI (left) and GT (right). Shading marks where the robust curve is ahead of the standard curve; dotted lines mark the 50\%, 80\%, and 90\% levels of Table~\ref{tab:attribution-concentration}. The robust curve reaches each threshold with fewer features (Section~\ref{sec:attribution-results}).}
\label{fig:attribution-concentration}
\end{figure}

Second, among active features, the robust SAE reaches each attribution-mass threshold with fewer of them in absolute terms: at 90\% attribution mass, IOI requires 276 features versus 343, and GT requires 63 versus 90 (Table~\ref{tab:attribution-concentration}, Figure~\ref{fig:attribution-concentration}). As a share of each model's pool the two are similar (IOI 23.1\% versus 21.3\%; GT 17.5\% versus 16.0\%), but a share does not measure how much must be inspected: the standard model's slightly smaller share of a larger pool is still more features (343 versus 276). GT probability difference is lower under adversarial training (Table~\ref{tab:qualification}), so fewer GT features can reflect a weaker behavior rather than cleaner features. We read the GT counts as consistent with IOI, not as an independent matched comparison. Zeroing each pool feature and measuring the resulting shift in the model's general output distribution (mean ablation-KL) shows the same directional pattern, though it is not seed-robust and one of three IOI seeds reverses direction (Appendix~\ref{app:selectivity-controls}).

\begin{table}[!ht]
\centering
\caption{Number of SAE features required to explain cumulative absolute target attribution. Reported values use the seed-42 SAE for each condition; the ordering holds at seeds 43 and 44 as well.}
\label{tab:attribution-concentration}
\small
\begin{tabular}{llrrrr}
\hline
Model & Task & Candidates & 50\% & 80\% & 90\% \\
\hline
Standard & IOI & 1,612 & 48 & 193 & 343 \\
Robust   & IOI & 1,196 & \textbf{43} & \textbf{163} & \textbf{276} \\
Standard & GT  & 563 & 14 & 54 & 90 \\
Robust   & GT  & 360 & \textbf{8} & \textbf{33} & \textbf{63} \\
\hline
\end{tabular}
\end{table}

\subsection{Circuit size on IOI depends on the faithfulness target}
\label{sec:circuit-results}

\begin{figure}[!ht]
\centering
\begin{minipage}[t]{0.49\textwidth}
\centering
\includegraphics[width=\linewidth]{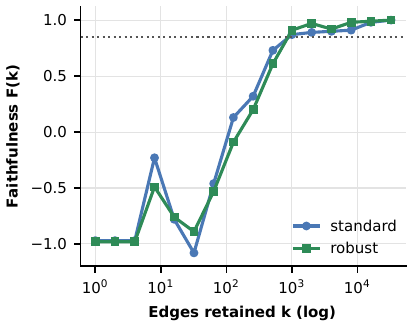}\\
\small\textbf{(a)}
\end{minipage}\hfill
\begin{minipage}[t]{0.49\textwidth}
\centering
\includegraphics[width=\linewidth]{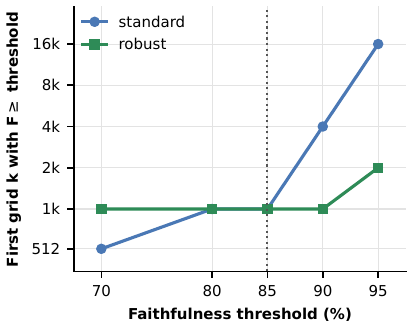}\\
\small\textbf{(b)}
\end{minipage}
\caption{Two readings of the same IOI edge-native faithfulness curve, EAP-IG, primary pair. \textbf{(a)} $F(k)$ against edges retained (log-$k$): the models cross 85\% at the same edge count, standard leads in the steep region ($k\approx64$--512), and net area favors standard (0.87 vs.\ 0.60). \textbf{(b)} First grid point reaching each threshold; a crossing at $k$ means the true crossing lies in $(k',k]$ for the preceding grid point $k'$. The dotted line marks 85\%. The grid is listed in Appendix~\ref{app:circuit-protocol}.}
\label{fig:tier3-edge}
\end{figure}

We recover faithful IOI circuits from the raw computational graph (156 nodes: 144 attention heads and 12 MLPs; 32,491 edges) using EAP-IG ($m=5$ interpolation steps, Section~\ref{sec:circuit-method}), on 200 prompts (100 ABBA / 100 BABA, name-swap counterfactuals). Table~\ref{tab:threshold-sensitivity} reports the first edge count reaching five faithfulness levels, rather than a single arbitrary threshold: at 70\% standard leads; at 80--85\% the models tie; at 90\% robust needs far fewer edges (1{,}000 versus 4{,}000); at 95\% the gap widens further (2{,}000 versus 16{,}000). Standard's curve plateaus after 85--90\%, requiring a much larger circuit to close the remaining gap to near-complete faithfulness; robust's curve continues climbing steeply over the same range (Figure~\ref{fig:tier3-edge}). Below 85\% faithfulness, neither subgraph reproduces the behavior well enough to constitute a circuit; the comparison there is between two non-faithful partial subgraphs, not between circuit sizes (raw values in Appendix~\ref{app:circuit-protocol}, Table~\ref{tab:fk-raw}). At and above 85\%, robust never needs more edges than standard, and needs substantially fewer at 90\% and 95\%. This high-faithfulness advantage is not specific to the primary pair: it replicates across the adversarial-strength sweep and a second corpus (Section~\ref{sec:generalization}, Table~\ref{tab:sweep-crossings}).

\begin{table}[!ht]
\centering
\caption{First edge count reaching each faithfulness threshold, IOI, EAP-IG, primary pair (\texttt{ft-clean} vs.\ \texttt{ft-e100-a05}). Both models are evaluated on the same geometric grid; a tie reflects grid resolution, not necessarily identical true crossing points. Left of the rule ($F<85\%$): partial subgraphs, not faithful circuits. Right of the rule ($F\geq85\%$): circuit-size comparison; 85\% is a tie.}
\label{tab:threshold-sensitivity}
\small
\begin{tabular}{lrr|rrr}
\hline
 & 70\% & 80\% & 85\% & 90\% & 95\% \\
\hline
Standard & 512 & 1{,}000 & 1{,}000 & 4{,}000 & 16{,}000 \\
Robust   & 1{,}000 & 1{,}000 & 1{,}000 & \textbf{1{,}000} & \textbf{2{,}000} \\
\hline
\end{tabular}
\end{table}

AUC (0.87 standard versus 0.60 robust) is the partial-subgraph recoverability estimand, as defined in Section~\ref{sec:circuit-method}; it does not speak to the edge count required to reach a faithful circuit at 90\% or 95\%. Conditional co-ablation, recovering backup components to account for self-repair at the node layer (Section~\ref{sec:circuit-method}), preserves the AUC ordering favoring standard at a much smaller scale (10--14 nodes, both models far below full faithfulness); it does not test the 90/95\% edge-level crossings above.

Greater-Than moves the same direction at 95\% (2{,}000 versus 8{,}000 edges). Because GT PD differs between models, we do not use that curve as a test of the main claim. Full detail, and the EAP-GP pilot, are in Appendix~\ref{app:circuit-protocol}.

\subsection{Representational and high-faithfulness circuit effects replicate across the sweep and a second corpus}
\label{sec:generalization}

The qualification sweep (Section~\ref{sec:qualification-results}) produced seven continually trained models spanning $\epsilon_{\mathrm{rel}}\in\{0.05,0.075,0.10\}$ and $\alpha\in\{0.2,0.5\}$, not just the single selected robust model used above. We extend all three axes to all seven, and additionally measure $\alpha$-ReQ, the power-law decay exponent of the residual-stream covariance spectrum (Appendix~\ref{app:alpha-req}), a representational-simplicity measure that does not depend on a trained SAE.

Across the sweep and FineWeb, SAE/$\alpha$-ReQ simplicity increases with robustness (Figure~\ref{fig:dose-response}; full values in Appendix~\ref{app:alpha-req}, Table~\ref{tab:dose-response}). The 85\% crossing is flat (1{,}000 edges for every model except \texttt{e050-a02}, which needs 2{,}000 and does not pass the masking gate), so circuit size at that threshold does not track the representational metrics. At higher thresholds it separates the models. For each of the four qualified robust models, the first edge count reaching 90\% faithfulness is 1{,}000--2{,}000, against 4{,}000 for the standard control, and the count reaching 95\% is 2{,}000, against 16{,}000 (Table~\ref{tab:sweep-crossings}). The 90\% crossing is ordered by $\alpha$ (1{,}000 edges at $\alpha=0.5$, 2{,}000 at $\alpha=0.2$); at 95\% all robust models sit at 2{,}000, so the grid does not separate them there. The two runs that fail the masking gate show the same direction. Because a reported crossing at $k$ means the true crossing lies in $(k',k]$, the robust and standard intervals are disjoint at both thresholds for every qualified robust model. The 90\% margin of the $\alpha=0.2$ model is one grid step; the 95\% margin is three.

\begin{table}[!ht]
\centering
\caption{First edge count reaching each faithfulness threshold, IOI, EAP-IG, for every model in the adversarial-strength sweep and the FineWeb pair, on the same 200-prompt protocol and geometric grid as Table~\ref{tab:threshold-sensitivity}. Left of the rule ($F<85\%$): partial subgraphs, not faithful circuits. Asterisks mark runs that fail the masking gate (Table~\ref{tab:qualification}). A reported count $k$ means the true crossing lies in $(k',k]$. At 90\% and 95\%, every robust model needs fewer edges than its matched standard control.}
\label{tab:sweep-crossings}
\small
\setlength{\tabcolsep}{4pt}
\begin{tabular}{lrrrr|rrr}
\hline
Model & $\epsilon_{\mathrm{rel}}$ & $\alpha$ & 70\% & 80\% & 85\% & 90\% & 95\% \\
\hline
\texttt{ft-clean} (standard)         & 0    & 0  & 512     & 1{,}000 & 1{,}000 & 4{,}000 & 16{,}000 \\
\texttt{e050-a02}$^*$                & .050 & .2 & 1{,}000 & 2{,}000 & 2{,}000 & 2{,}000 & 2{,}000 \\
\texttt{e075-a02}$^*$                & .075 & .2 & 1{,}000 & 1{,}000 & 1{,}000 & 2{,}000 & 2{,}000 \\
\texttt{e100-a02}                    & .100 & .2 & 1{,}000 & 1{,}000 & 1{,}000 & 2{,}000 & 2{,}000 \\
\texttt{e050-a05}                    & .050 & .5 & 1{,}000 & 1{,}000 & 1{,}000 & 1{,}000 & 2{,}000 \\
\texttt{e075-a05}                    & .075 & .5 & 512     & 1{,}000 & 1{,}000 & 1{,}000 & 2{,}000 \\
\textbf{\texttt{e100-a05}} (primary) & \textbf{.100} & \textbf{.5} & 1{,}000 & 1{,}000 & 1{,}000 & \textbf{1{,}000} & \textbf{2{,}000} \\
\hline
\texttt{fw-clean} (standard, FineWeb) & 0    & 0  & 1{,}000 & 1{,}000 & 1{,}000 & 4{,}000 & 8{,}000 \\
\texttt{fw-robust} (FineWeb)          & .100 & .5 & 512     & 1{,}000 & 1{,}000 & \textbf{1{,}000} & \textbf{2{,}000} \\
\hline
\end{tabular}
\end{table}

\begin{figure}[!ht]
\centering
\includegraphics[width=\textwidth]{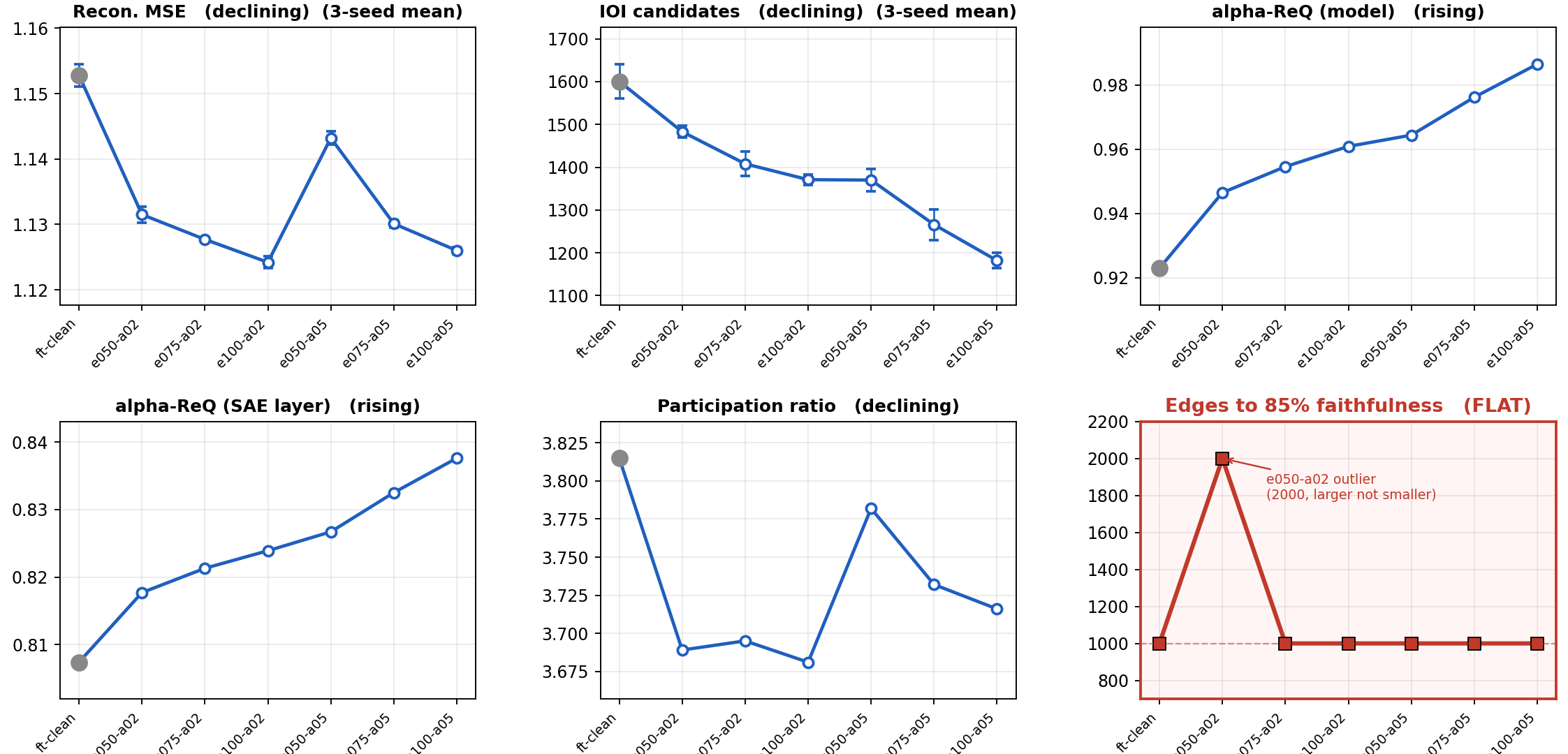}
\caption{Representational metrics (panels 1--5) move in the predicted direction with adversarial strength across the seven-model sweep, ordered by measured $\Delta_{\mathrm{adv}}$ (1.642 $\to$ 0.362); two adjacent configurations with near-identical measured robustness (\texttt{e100-a02}, \texttt{e050-a05}) produce small non-monotonic steps in reconstruction MSE and participation ratio. Panel 6 shows the single 85\% qualification gate, flat at 1{,}000 edges except for one outlier (\texttt{e050-a02}, 2{,}000 edges -- larger, not smaller). Crossings at 90\% and 95\% are in Table~\ref{tab:sweep-crossings}. Reconstruction MSE (inference-threshold mode) and IOI candidates are three-seed means with $\pm 1$ SD error bars (seeds 42/43/44); $\alpha$-ReQ, participation ratio, and edges-to-85\% are single deterministic measurements per model (Appendix~\ref{app:alpha-req}).}
\label{fig:dose-response}
\end{figure}

We separately test whether the representational trend depends on OpenWebText specifically, using one additional standard-robust pair continually trained identically to the primary comparison except for the pretraining corpus (FineWeb in place of OpenWebText; Appendix~\ref{app:corpus-ablation}). Every representational metric moves in the same direction as the primary result, and the 85\% edge count is again flat; at 90\% and 95\% the robust model again needs fewer edges (1{,}000 versus 4{,}000 and 2{,}000 versus 8{,}000; Appendix~\ref{app:corpus-ablation}, Table~\ref{tab:corpus-ablation}).

\section{Conclusion}
\label{sec:discussion}

In this paper, we test whether adversarial training reduces reverse-engineering complexity, using it as a controlled instrument that reshapes representations while task competence is held fixed. To this end, we compare matched standard and adversarially trained GPT-2 Small models along three complementary axes: SAE decomposability, SAE feature engagement, and the size of faithful circuits recovered from the raw computational graph.

Our results reveal three key findings: (1) the robust model is more SAE-decomposable and engages fewer active SAE features, corroborated by $\alpha$-ReQ, an SAE-free measure that rules out an SAE-specific artifact; (2) faithful circuit size is regime-dependent -- standard leads or ties below 85\% faithfulness, but robust needs substantially fewer edges at high faithfulness (90\%, 95\%); and (3) both the representational trend and the high-faithfulness circuit advantage replicate across a seven-point adversarial-strength sweep (all four qualified robust models) and a second corpus. These results concern one training instrument (embedding-space PGD) on GPT-2 Small, with IOI as the only competence-matched task; whether the pattern extends to other attacks, tasks, architectures, and scales remains open, and we regard this work as a starting point for testing it. Our broader motivation is to test whether robustness can improve model auditing and reshape how interpretability-oriented training is prioritized; this work is a first step toward that goal.

\section{Related Work}
\label{sec:related-work}

Adversarial examples can exploit predictive but non-robust features, and adversarial training changes which features a model uses \citep{tsipras2019robustness,ilyas2019adversarial}, with robustness linked to more human-aligned gradients, saliency maps, and representations \citep{ross2017improvingadversarialrobustnessinterpretability,etmann2019connectionadversarialrobustnesssaliency,engstrom2019adversarialrobustnesspriorlearned,wang2022robust}; these results concern input sensitivity or representational geometry, not causal mechanism size. A closer precedent comes from vision: adversarially trained networks admit sparser attribution vectors \citep{chalasani2020concise} and exhibit feature purification \citep{allenzhu2020featurepurification}, with gradients and saliency maps aligning more closely with human-interpretable structure \citep{kim2019bridgingadversarialrobustnessgradient} -- predicting simpler features and sparser attributions, not a smaller causal graph, the prediction we test.

Superposition offers a sharper hypothesis, that models represent more features than dimensions by accepting interference \citep{elhage2022superposition}, recently linked to adversarial vulnerability \citep{gorton2025adversarialexamplesbugssuperposition,stevinson2026adversarialattacksleverageinterference,gao2026featurecompression,zhang2024interpretabilitygainsfeaturemonosemanticity,bereska2025superpositionlossycompressionmeasure}; SAEs measure this decomposition at scale \citep{huben2024sparse,gao2025scaling}, though reconstruction and sparsity alone do not guarantee causal features \citep{karvonen2025saebench,paulo2026different,li2026evaluatingadversarialrobustnessconcept}, motivating our separate evaluation of decomposability and causal structure. On the causal side, manual and automated methods have identified and scaled circuit recovery for GPT-2 Small behaviors \citep{wang2022interpretability,hanna2023greaterthan,conmy2023acdc,syed2024attribution,hanna2024have}, though recovered circuits remain conditional on the attribution method, threshold, and pruning procedure \citep{hanna2024have}, and ablations can trigger self-repair that understates causal dependence \citep{mcgrath2023hydra,rushing2024selfrepair}.

The closest empirical studies apply mechanistic interpretability to adversarial vulnerability: \citet{garciacarrasco2024vulnerabilities} localize components affected by adversarial examples in a fixed GPT-2 Small circuit, and \citet{walkowiak2025unpackingrobustnessinflectionallanguages} combine adversarial evaluation with EAP-IG circuits across languages; \citet{bereska2024robust} lists reverse-engineering robust models in a self-published proposal, not executed empirically. To our knowledge, we are the first to compare matched standard and adversarially trained language models through SAE decomposability, SAE feature engagement, and the size of established task circuits at fixed faithfulness. Appendix~\ref{app:extended-related-work} discusses SAE evaluation, circuit-discovery variants, and self-repair in greater detail.

\section*{Reproducibility Statement}

We will publicly release our code, trained checkpoints, and SAEs. Complete continual-training and attack specifications are in Appendix~\ref{app:training-details}; data partitions, prompt generation, and provenance in Appendix~\ref{app:data-provenance}; SAE training and per-seed statistics in Appendix~\ref{app:sae-details}; attribution and selectivity controls in Appendix~\ref{app:selectivity-controls}; and the circuit-recovery protocol, edge grid, and raw $F(k)$ values in Appendix~\ref{app:circuit-protocol}. Seeds and the implementation's test coverage are described in Appendix~\ref{app:reproducibility}.

\section*{AI Use Disclosure}

We used Claude (Anthropic) to assist with manuscript preparation.

\textbf{Writing and polishing.} Claude assisted with proofreading, polishing, and structuring the paper's sections.

\textbf{Citations.} Claude assisted with verifying citations against primary sources and identifying missing bibliography entries.

\textbf{LaTeX and formatting.} Claude assisted with LaTeX formatting, table and figure layout, and troubleshooting page-length and float-placement issues.

All experimental design, model training, data analysis, and interpretation of results were carried out by the authors. All claims, figures, and interpretations were reviewed and approved by the authors, who take full responsibility for the submission's content and accuracy.

\nocite{malik2023extreme}
\bibliography{iclr2027_conference}
\bibliographystyle{iclr2027_conference}

\clearpage
\appendix

\section{Extended Related Work and Methodological Context}
\label{app:extended-related-work}

\paragraph{Conventional interpretations of robust models.}
The connection between robustness and interpretability has most often been studied through gradients and visual representations.
Input-gradient regularization can improve both resistance to transferred adversarial examples and the human legibility of gradients \citep{ross2017improvingadversarialrobustnessinterpretability}.
Robustness has also been connected to the alignment of saliency maps with perceptually relevant directions \citep{etmann2019connectionadversarialrobustnesssaliency,kim2019bridgingadversarialrobustnessgradient}, while robust image classifiers learn representations and attributions that are often more human-aligned \citep{engstrom2019adversarialrobustnesspriorlearned,wang2022robust}.
Conversely, training for robust interpretation can support robust classification \citep{boopathy2020proper,noack2020empiricalstudyrelationnetwork}.
These findings provide evidence for a relationship between robustness and conventional interpretability, but each operationalizes interpretability through input-output sensitivity, attribution, or representation quality rather than through the causal organization of a learned computation.

\paragraph{Evaluating sparse autoencoders.}
SAEs can recover fine-grained language-model features and scale to large overcomplete dictionaries \citep{huben2024sparse,gao2025scaling}.
Their evaluation nevertheless remains unsettled.
SAEBench finds that improvements on reconstruction and sparsity proxies do not reliably transfer to practical interpretability tasks \citep{karvonen2025saebench}, while semantic evaluations based on polysemous words expose failures hidden by the usual reconstruction-sparsity frontier \citep{minegishi2025rethinking}.
Independent SAEs trained on the same model and data can learn substantially different dictionaries \citep{paulo2026different}, motivating explicit measurement of feature consistency across runs \citep{song2026consistency}.
A recent preprint further reports high noise or weak discriminability in several benchmark metrics \citep{chanin2026reliable}.
SAE representations can themselves be adversarially fragile: small input perturbations can manipulate concept interpretations without materially changing the underlying language-model activations \citep{li2026evaluatingadversarialrobustnessconcept}.
These limitations do not make SAEs uninformative, but they prevent reconstruction, sparsity, or automated feature labels from serving as stand-alone evidence of simpler computation.

\paragraph{Automated circuit discovery.}
Automated Circuit Discovery (ACDC) recursively removes edges using activation-patching effects and recovers compact subgraphs for established tasks \citep{conmy2023acdc}.
Edge Attribution Patching (EAP) replaces repeated interventions with a gradient-based approximation, reducing circuit discovery to two forward passes and one backward pass \citep{syed2024attribution}.
EAP with integrated gradients improves faithfulness relative to local-gradient EAP and demonstrates that overlap with a reference circuit can remain high even when the recovered circuit is behaviorally unfaithful \citep{hanna2024have}.
EAP-GP modifies the integration path to mitigate saturation and further improve gradient-based circuit identification \citep{zhang2025eapgp}.
Sparse feature circuits extend causal graph discovery from attention heads and MLPs to SAE features \citep{marks2025sparsefeaturecircuits}, while recent formal work defines circuit guarantees relative to explicit input domains and intervention semantics \citep{hadad2026formal}.
Together, these methods make circuit recovery scalable, but their outputs remain conditional on the graph granularity, patching baseline, task metric, attribution estimator, threshold, and search procedure.
Our circuit-size comparison therefore concerns the smallest graph recovered by a fixed algorithm at a specified faithfulness threshold, not a global minimum over every possible circuit.

\paragraph{Circuit stability, selectivity, and self-repair.}
Circuit analyses can identify similar functional algorithms across training checkpoints and model scales even when individual components change \citep{tigges2024consistent}.
Conversely, attention-pattern selectivity need not identify the components that causally implement a task \citep{xu2026pattern}.
These results reinforce the distinction between observational structure and causal dependence.
Causal interventions introduce their own complication: ablating a component can change the behavior of downstream components, producing Hydra-like compensation \citep{mcgrath2023hydra}.
Self-repair occurs across language-model families and scales and can arise through several mechanisms, including LayerNorm scaling and reduced erasure of upstream contributions \citep{rushing2024selfrepair}.
Copy suppression provides a detailed example in which an attention-head motif explains both a component's normal function and part of the compensation observed after ablation \citep{mcdougall2024copysuppression}.
Because one-at-a-time ablations can therefore underestimate distributed dependence, our causal analysis includes conditional co-ablation to test whether apparently weak components become important when potential substitutes are simultaneously removed.

\paragraph{Relation to adversarial mechanistic interpretability.}
\citet{garciacarrasco2024vulnerabilities} propose a pipeline that first identifies a task circuit in a fixed GPT-2 Small model, then generates adversarial samples and localizes the affected components.
\citet{walkowiak2025unpackingrobustnessinflectionallanguages} combine adversarial attacks with EAP-IG circuits to compare language and inflectional settings.
The feature-level literature instead studies whether superposition or representation compression creates adversarial vulnerability \citep{gorton2025adversarialexamplesbugssuperposition,bereska2025superpositionlossycompressionmeasure,stevinson2026adversarialattacksleverageinterference,gao2026featurecompression}.
Finally, \citet{bereska2024robust} lists reverse-engineering robust models as an explicit research objective in a self-published proposal, alongside investigating feature superposition and designing training methods that improve both robustness and interpretability; the proposal is not executed empirically.
Our work carries out that objective: we experimentally compare matched standard and adversarially trained language models across three non-equivalent levels: sparse feature decomposition, SAE feature engagement, and fixed-faithfulness circuit recovery.

\section{Continual-training and attack details}
\label{app:training-details}

\subsection{Initialization and optimization}

Every run initializes from \texttt{openai-community/gpt2}, revision \texttt{607a30d783dfa663caf39e06633721c8d4cfcd7e}. The checkpoint has 124M parameters, a 50,257-token vocabulary, and learned affine biases. We update all parameters rather than using parameter-efficient adapters. The training stream is drawn from the Hugging Face parquet conversion of OpenWebText at revision \texttt{433fe0f44ed7894fea29c08b3202aa348ccc6369}. The standard and adversarial runs receive the same examples in the same order.

\begin{table}[h]
\centering
\caption{Continual-training hyperparameters shared by all seven runs.}
\label{tab:training-hyperparameters}
\small
\begin{tabular}{ll}
\hline
Hyperparameter & Value \\
\hline
Training budget & exactly 1,000,000,000 tokens \\
Updates & 1,907 full updates and one final partial update \\
Global batch & 524,288 tokens per update \\
Micro-batch & 16 sequences, with gradient accumulation \\
Optimizer & AdamW \\
Adam coefficients & $(\beta_1,\beta_2)=(0.9,0.95)$ \\
Peak learning rate & $5\times10^{-5}$ \\
Schedule & 100-step warmup, cosine decay to $5\times10^{-6}$ \\
Weight decay & 0.1 \\
Gradient clipping & global norm 1.0 \\
Precision & bfloat16 autocast \\
Checkpoint interval & every 250 updates \\
\hline
\end{tabular}
\end{table}

The attack operates on the output of the token-embedding lookup before positional embeddings are added. In each micro-batch, its absolute radius is
\begin{equation}
    \epsilon_{\mathrm{abs}}
    = \epsilon_{\mathrm{rel}}
      \frac{1}{|\mathcal{I}|}
      \sum_{t\in\mathcal{I}}\|E(x_t)\|_2,
\end{equation}
where $\mathcal{I}$ contains non-special-token positions. Each token perturbation is initialized at a random point inside its $\ell_2$ ball. We then perform ten normalized-gradient ascent steps of size $2.5\epsilon_{\mathrm{abs}}/10$, projecting every token independently after each step. Token 50256 is excluded from perturbation. Training and qualification use the same attack implementation and parameterization.

\subsection{Why continual training is used}
\label{app:preliminary-controls}

We initially attempted a matched comparison trained from scratch on 9.9B FineWeb tokens. The standard model reached only 55.4\% IOI accuracy, compared with 99.7\% for the official GPT-2 checkpoint on the same 1,000 prompts. Thus, the target behavior did not reliably emerge within the available from-scratch budget. A one-step embedding-adversarial variant also showed catastrophic overfitting: its loss was 1.20 under its training-time attack but 24.70 under PGD-10, a 23.5-nat discrepancy. These failures motivated continual training from a checkpoint that already possesses the target computations and the use of multi-step PGD during both training and verification.

\section{Evaluation data and provenance}
\label{app:data-provenance}

Table~\ref{tab:data-partitions} summarizes the OpenWebText partitions. Model-training data, robustness evaluation, SAE training, SAE evaluation, and the generic-language-modeling attribution control are positionally separated in the token stream. Because this OpenWebText conversion has no native document identifiers, we synthesize identifiers from parquet file and row indices and record hashes of the resulting lists. One document crosses each of two token-level partition boundaries; the token streams themselves remain disjoint.

\begin{table}[h]
\centering
\caption{OpenWebText partitions used by the study.}
\label{tab:data-partitions}
\small
\begin{tabular}{lrl}
\hline
Partition & Tokens & Provenance note \\
\hline
Continual training & 1,000,000,000 & 973,424 logged document IDs \\
Robustness evaluation & 100,000,000 & 88,637 reconstructed IDs \\
SAE training & 300,000,000 & post-training stream \\
SAE evaluation & 10,000,000 & subsequent held-out stream \\
Attribution control & 256,000 & 1,000 sequences of length 256 \\
\hline
\end{tabular}
\end{table}

The IOI set contains 1,000 prompts, split evenly between ABBA and BABA templates, generated with the official IOIDataset implementation at seed 42. Names are restricted to single GPT-2 tokens. The GT set contains 1,000 balanced prompts generated with the official YearDataset implementation at seed 42. To construct GT counterfactuals, we use the official ``01-dataset'' bad-sentence branch and require matching start-year token counts; 973 prompts pass this parity filter. The unused bad-sentence branch of the original generator was patched to avoid a batching crash when a sampled counterfactual year is represented by one BPE token. The sampling procedure, templates, and random-number stream are otherwise unchanged.

\section{Qualification sweep and behavioral trade-off}
\label{app:qualification-sweep}

Qualification metrics are computed for every sweep cell rather than stopping after the first passing candidate. All adversarial runs clear the IOI retention floor, while four satisfy the masking sanity check. The standard model's large single-step versus PGD-10 gap is reported as a non-robust reference and is not interpreted as a failed adversarial-training run.

The GT results reveal a graded behavioral trade-off that IOI accuracy does not expose. Paired prompt-level differences from the standard model are $-0.063\pm0.005$, $-0.087\pm0.006$, $-0.117\pm0.007$, $-0.112\pm0.006$, $-0.171\pm0.008$, and $-0.229\pm0.010$ for \texttt{e050-a02}, \texttt{e075-a02}, \texttt{e100-a02}, \texttt{e050-a05}, \texttt{e075-a05}, and \texttt{e100-a05}, respectively. The original checkpoint scores 0.706 on this prompt set, the continually trained standard model scores 0.671, and the selected robust model scores 0.442. Because prompt sampling differs from the setup used to obtain previously published headline values, the paired within-set difference is the primary GT comparison.

\section{Sparse-autoencoder implementation and statistics}
\label{app:sae-details}

We cache block-8 residual activations once per model as bfloat16 shards, recording a hash of the corresponding model weights. Each SAE is trained for 73,242 updates, one pass over 300M tokens, with batches of 4,096 tokens and Adam at learning rate $4\times10^{-4}$ and $(\beta_1,\beta_2)=(0.9,0.99)$. Training uses a roughly 500,000-token activation buffer, cross-shard shuffling, and refill when half the buffer has been consumed. No early stopping is used.

Inputs are normalized per token by $s=\sqrt{d}/\|x\|_2$ before encoding and returned to the original scale after decoding. BatchTopK selects $32$ times the number of batch tokens activations across the batch. Decoder columns are constrained to unit norm by radially projecting their gradients before each update and renormalizing after it. AuxK uses 512 features, coefficient $1/32$, and defines a feature as dead if it has not been selected for 10M tokens. At most one dead feature per seed (Table~\ref{tab:sae-per-seed}).

For downstream inference, batch-relative selection is replaced by a fixed threshold. The threshold is the mean, across held-out evaluation batches, of the smallest positive activation selected by BatchTopK. At seed 42, the thresholds are 0.785 for the standard SAE and 0.741 for the robust SAE. This prevents the activation state of one prompt from depending on other prompts in the evaluation batch.

We quantify measurement uncertainty by pairing residual positions between models and clustering observations into 1,024-token sequence blocks. Across 9,756 blocks, standard-minus-robust reconstruction MSE is $0.0323\pm0.0007$ in original space and $0.0112\pm0.0001$ in normalized space. We separately retrain each SAE with seeds 42, 43, and 44. The original-space MSE ranges do not overlap: $1.1601\pm0.0022$ for the standard activations and $1.1293\pm0.0009$ for the robust activations. Reconstruction statistics, SAE feature engagement (Section~\ref{sec:attribution-results}), and the ablation-KL population comparison (Appendix~\ref{app:selectivity-controls}) use all three seeds.

Table~\ref{tab:sae-per-seed} reports the per-seed values underlying the aggregate statistics above.

\begin{table}[h]
\centering
\caption{Per-seed SAE statistics. Bold values indicate the robust condition.}
\label{tab:sae-per-seed}
\small
\begin{tabular}{lrrrrrr}
\hline
Metric & clean s42 & clean s43 & clean s44 & robust s42 & robust s43 & robust s44 \\
\hline
MSE, original residual space (train-mode) & 1.1624 & 1.1579 & 1.1601 & \textbf{1.1301} & \textbf{1.1295} & \textbf{1.1284} \\
MSE, inference-threshold mode & 1.1546 & 1.1511 & 1.1526 & \textbf{1.1262} & \textbf{1.1265} & \textbf{1.1254} \\
Final training loss (normalized space) & 0.0877 & 0.0902 & 0.0869 & \textbf{0.0768} & \textbf{0.0770} & \textbf{0.0788} \\
Dead latents & 0 & 1 & 0 & 0 & 0 & 1 \\
Inference threshold $T$ & 0.785 & 0.783 & 0.787 & \textbf{0.741} & \textbf{0.742} & \textbf{0.743} \\
\hline
\end{tabular}
\end{table}

Table~\ref{tab:concentration-per-seed} reports per-seed candidate counts and feature counts at the 90\% attribution-mass threshold underlying Table~\ref{tab:attribution-concentration} and Section~\ref{sec:attribution-results}. The robust SAE requires fewer candidates and fewer features at every seed, both tasks, with no exceptions.

\begin{table}[h]
\centering
\caption{Per-seed SAE feature engagement. Values are candidate count / features for 90\% attribution mass, per seed.}
\label{tab:concentration-per-seed}
\small
\begin{tabular}{lll}
\hline
& \texttt{ft-clean} (s42/s43/s44) & \texttt{ft-e100-a05} robust (s42/s43/s44) \\
\hline
GT candidate features & 563 / 561 / 533 & \textbf{360 / 385 / 357} \\
GT features for 90\% & 90 / 78 / 85 & \textbf{63 / 61 / 59} \\
IOI candidate features & 1,612 / 1,634 / 1,556 & \textbf{1,196 / 1,188 / 1,162} \\
IOI features for 90\% & 343 / 363 / 341 & \textbf{276 / 280 / 262} \\
\hline
\end{tabular}
\end{table}

\section{Attribution implementation and selectivity controls}
\label{app:selectivity-controls}

For IOI, target attribution is summed over the first subject, indirect object, second subject, and final readout positions. For GT, it is summed over the start-year span and final readout. A feature enters the candidate set if it activates at any intervention position on a clean or corrupted prompt. Each model-task analysis ranks its own candidate features; rankings are never merged across SAEs. The ablation pool includes all features tied at the top-100 boundary.

\subsection{Non-discriminative generic-language-modeling control}

On held-out OpenWebText, generic next-token attributions are $\sim 10^{-8}$ against task attributions $\sim 10^{-2}$. A selectivity ratio against that control is uninformative: the denominator is negligible in both models.

\subsection{Ablation-KL population comparison}

We zero-ablate each of the top-100 most-attributed features and measure $D_{\mathrm{KL}}(p\|p_i^{\mathrm{abl}})$ from the unspliced to the ablated output distribution, using a real intervention rather than a gradient-based estimate: KL divergence has zero first derivative at the unablated distribution, so a linearized (attribution-patching) estimate of this quantity is identically zero regardless of the true effect. For GT, ablation is applied at the readout position, where pool features are already reliably active. For IOI, an earlier readout-only version left 98 of the top 100 features permanently inactive, since IOI features act predominantly at the name positions; we instead ablate at the union of positions in $\mathcal{T}_{\mathrm{IOI}}$ where a given feature is active on that prompt, which resolves this and leaves all 100 features active for both models.

Because the two SAEs are trained independently, a feature index carries no correspondence across models, so we do not pair features by index or form a per-feature ratio against target attribution. Instead, for each model and task we treat the resulting 100 ablation-KL values as an unpaired population and compare the two models' populations directly, across all three SAE seeds. Table~\ref{tab:kl-population} reports the full per-seed detail. Values exclude control-set-dead features (zero effect by construction, since ablating an already-inactive feature is a no-op); dead counts are reported separately per cell. Medians are close between models on both tasks; the separation is concentrated in the 90th-percentile tail. We report this as a directional pattern only: with three seeds, per-seed means are not separated by standard deviation, and one of three IOI seeds reverses direction.

\begin{table}[h]
\centering
\caption{Ablation-KL effect, robust vs. standard, excluding control-set-dead features. Mean/median in units of $10^{-3}$.}
\label{tab:kl-population}
\small
\begin{tabular}{llrrrr}
\hline
Task & Model & Dead (s42/43/44) & Mean (excl.) & Median (excl.) & p90 (excl.) \\
\hline
GT & Standard & 14/16/15 & 7.4 / 8.3 / 10.7 & $\sim$1.0 & 10--14 \\
GT & Robust   & 16/16/21 & \textbf{7.0 / 6.0 / 9.3} & \textbf{0.6--1.1} & \textbf{10--12} \\
IOI & Standard & 0/0/0 & 5.4 / 8.5 / 10.9 & $\sim$0.9--1.0 & 18--20 \\
IOI & Robust   & 0/0/0 & 6.4 / 6.2 / 7.2 & $\sim$0.9 & \textbf{9--15} \\
\hline
\end{tabular}
\end{table}

\section{Circuit recovery protocol}
\label{app:circuit-protocol}

For each model-task pair, we score all 32,491 edges of the 156-node raw computational graph (144 attention heads, 12 MLPs) using EAP-IG ($m=5$ interpolation steps) and, as a pilot, EAP-GP, on identical prompt sets and counterfactuals: 200 IOI prompts (100 ABBA / 100 BABA, name-swap counterfactuals) and 200 of the first 207 GT prompts scanned, surviving the start-year parity filter (collection stops at 200 passes, so this count is smaller than the full-set 973-of-1000 in Appendix~\ref{app:data-provenance}), "01-dataset" counterfactuals. Circuit size is the smallest edge count, from a geometric grid $k\in\{1,2,4,8,16,32,64,128,256,512,1000,2000,4000,8000,16000,32491\}$ (a denser tail past 512, not a continued power-of-two sequence, to avoid skipping the 85\% crossing), that reaches a given fraction of full-graph faithfulness (85\% as the qualification gate; 70--95\% in Table~\ref{tab:threshold-sensitivity}; all five thresholds for the full sweep in Table~\ref{tab:sweep-crossings}), together with the trapezoidal area under the full faithfulness curve on a log-$k$ axis as a finer-grained discriminator.

Table~\ref{tab:fk-raw} reports the raw $F(k)$ values underlying Figure~\ref{fig:tier3-edge}, IOI, EAP-IG, per Equation~\ref{eq:faithfulness}. Below $k=1000$, several entries are negative, meaning that subgraph performs worse than the fully corrupted baseline; at $k=128$ standard leads by its largest margin in the curve ($F=0.13$ versus $-0.09$ robust), still well short of the 85\% threshold. Negative values at small $k$ likely reflect $|\mathrm{score}|$ ranking admitting edges with negative effect; signed ranking is left to future work.

\begin{table}[h]
\centering
\caption{Raw $F(k)$, EAP-IG. IOI underlies Figure~\ref{fig:tier3-edge}. GT underlies the 95\% sentence in Section~\ref{sec:circuit-results}; GT is not competence-matched.}
\label{tab:fk-raw}
\small
\begin{tabular}{lrrrrrrrr}
\hline
$k$ & 1 & 2 & 4 & 8 & 16 & 32 & 64 & 128 \\
\hline
IOI Standard & $-0.97$ & $-0.97$ & $-0.97$ & $-0.23$ & $-0.78$ & $-1.08$ & $-0.46$ & $0.13$ \\
IOI Robust   & $-0.98$ & $-0.98$ & $-0.98$ & $-0.49$ & $-0.76$ & $-0.89$ & $-0.53$ & $-0.09$ \\
GT Standard  & 0.00 & 0.00 & 0.00 & 0.00 & 0.15 & 0.38 & 0.66 & 0.88 \\
GT Robust    & 0.00 & 0.00 & 0.00 & 0.07 & 0.14 & 0.31 & 0.65 & 0.88 \\
\hline
\end{tabular}

\begin{tabular}{lrrrrrrrr}
\hline
$k$ & 256 & 512 & 1000 & 2000 & 4000 & 8000 & 16000 & 32491 \\
\hline
IOI Standard & 0.32 & 0.73 & 0.87 & 0.89 & 0.90 & 0.91 & 0.98 & 1.00 \\
IOI Robust   & 0.20 & 0.61 & 0.91 & 0.97 & 0.92 & 0.98 & 0.99 & 1.00 \\
GT Standard  & 0.92 & 0.89 & 0.93 & 0.94 & 0.87 & 0.97 & 0.97 & 1.00 \\
GT Robust    & 0.94 & 0.94 & 0.89 & 0.99 & 0.99 & 0.96 & 1.00 & 1.00 \\
\hline
\end{tabular}
\end{table}

GT-standard's first grid point reaching 95\% is $k=8000$: $k=2000$ (0.94) and $k=4000$ (0.87, non-monotonic) both fall short.

EAP-GP is a recently proposed variant using integrated gradient paths intended to mitigate the saturation effect known to affect EAP-IG. No public reference implementation existed at the time of this work; our implementation follows the method description, with one identified approximation -- a single shared gradient path computed at the input activations and reused for all downstream edges, rather than one path per node, matching the roughly 5$\times$ per-step cost the source paper reports relative to EAP-IG. This approximation is not independently validated against the source paper's own reported results. EAP-GP did not outperform EAP-IG on our graph for either task or model: area under the faithfulness curve was negative for EAP-GP on IOI for both models (standard: $-0.97$; robust: $-0.70$), against positive AUC for EAP-IG (standard: 0.87; robust: 0.60). We select the primary method per task by comparing AUC between EAP-IG and EAP-GP before inspecting the standard-versus-robust comparison, and report EAP-IG throughout the main text on this basis. We do not draw conclusions about EAP-GP's general effectiveness from this result; a genuine implementation bug, rather than a property of the method on this graph, cannot be ruled out.

\begin{figure}[h]
\centering
\includegraphics[width=0.75\textwidth]{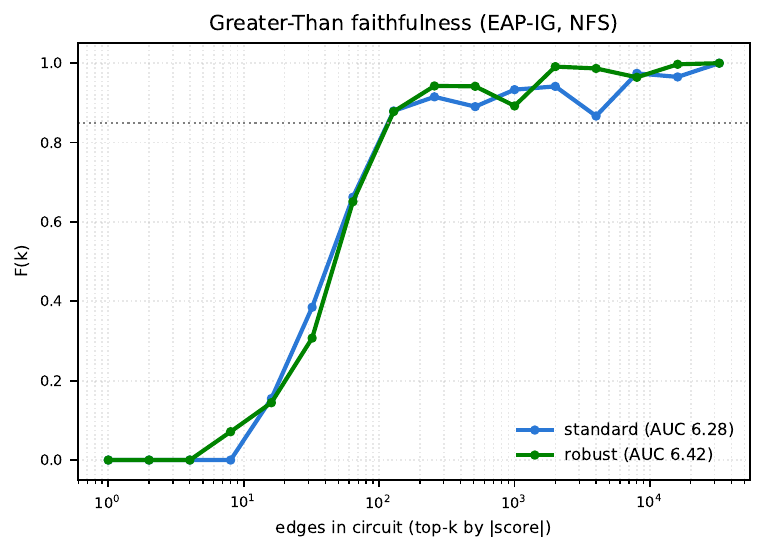}
\caption{Greater-Than edge-native faithfulness, $F(k)$, EAP-IG (normalized-faithfulness-score, or NFS, normalization). AUC is an unnormalized trapezoid of $F(k)$ over $\ln k$ (maximum $\approx10.4$ if $F=1$ throughout the grid); GT's curve stays near zero or above and rises early, while IOI's curve dips to $F\approx-1.08$ before climbing (Table~\ref{tab:fk-raw}), which is why GT and IOI AUCs sit on different numeric scales and are not directly comparable. Not competence-matched between models; not used as evidence for the main result (Section~\ref{sec:circuit-results}).}
\label{fig:tier3-gt-eapig}
\end{figure}

\begin{figure}[h]
\centering
\includegraphics[width=\textwidth]{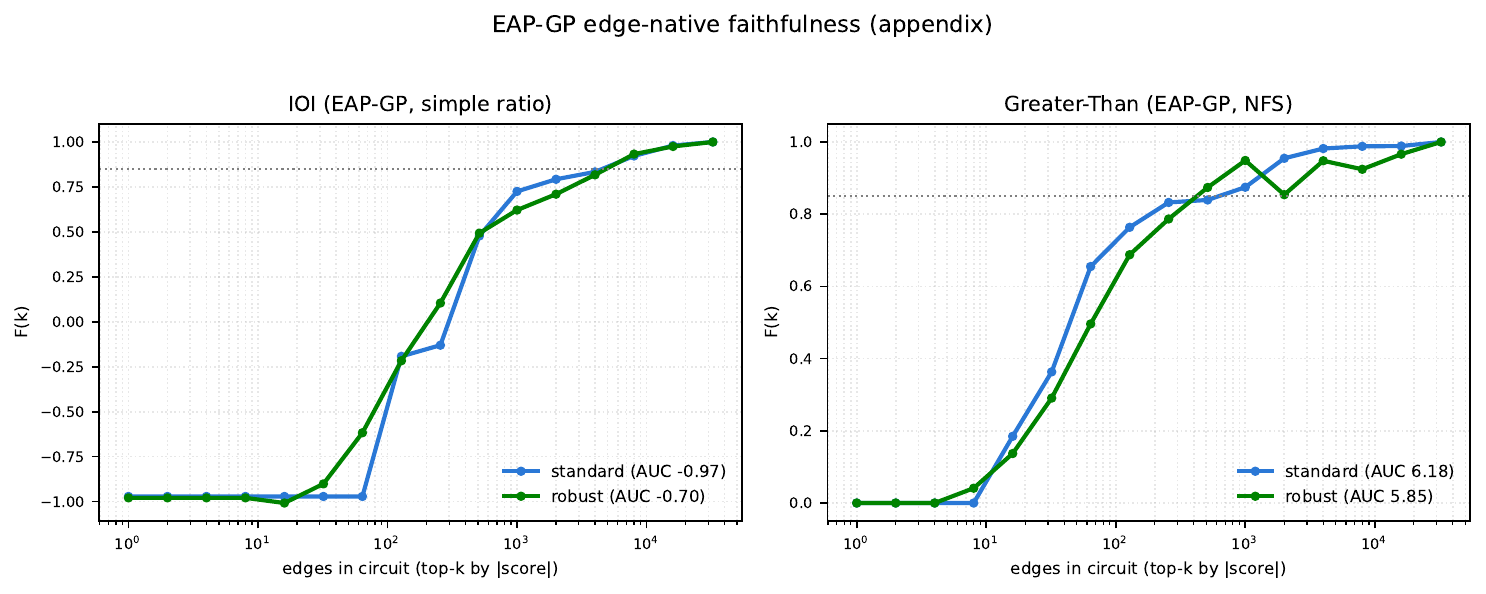}
\caption{EAP-GP edge-native faithfulness, both tasks. Negative AUC on IOI for both models; EAP-IG is used for the headline result on this basis.}
\label{fig:tier3-eapgp}
\end{figure}

Top-$k$ edge selection can yield a disconnected subgraph, which could in principle make a circuit look less faithful for reasons unrelated to the underlying attribution method's quality. As a check, we rebuilt circuits with a connectivity-preserving greedy edge selector in place of plain top-$k$ and recomputed both methods' faithfulness curves under it. The verdict does not change: EAP-IG beats EAP-GP under both builders, on both tasks and models. The greedy builder improves both methods' AUC, since connected circuits are generally more faithful, but it improves EAP-IG more than EAP-GP on IOI (+0.81 versus +0.48), so EAP-IG's advantage widens rather than narrows under the connectivity-preserving builder. The same holds on Greater-Than, our secondary check: EAP-IG's margin over EAP-GP moves from $+0.10$/$+0.56$ (standard/robust, top-$k$) to $+0.13$/$+0.54$ (greedy), essentially unchanged. Top-$k$ disconnection is therefore not the source of EAP-GP's underperformance on either task.

Conditional co-ablation (CoAx) recovers backup components after primary-circuit ablation, following the scoring formula of \citet{gong2026coax}\footnote{Propositions 2--3.}, and tests whether apparently weak components become necessary once substitutes are removed, guarding against self-repair understating causal dependence. This step costs a fixed $295$ forward passes for backup recovery plus $22$ for the backup-validity gate, per model, task, and attribution method -- constant by construction (a function of the candidate set size, not of which circuit is being recovered), so it does not itself discriminate between the standard and robust conditions; the resulting circuit's faithfulness curve does. Node-level faithfulness on the primary-plus-backup set remains far below 85\% for both models (10--14 nodes). It is not a second 90/95\% edge-level test.

\section{$\alpha$-ReQ: representation eigenspectrum decay}
\label{app:alpha-req}

$\alpha$-ReQ is a representational-simplicity measure independent of any trained SAE, used to test whether the dose-response in Section~\ref{sec:generalization} is an artifact of the SAE family rather than a property of the underlying representation. For each residual-stream layer (every block's \texttt{resid\_post}, plus the layer entering block 8 used for the SAE), we accumulate the centered activation covariance over a fixed evaluation token set, take its eigenvalues $\lambda_1\geq\ldots\geq\lambda_d$ ($d=768$), and fit $\lambda_i\sim i^{-\alpha}$ by OLS on a log-log scale over ranks 6--691 (dropping the top-5 spikes, which reflect a small number of dominant directions rather than the bulk spectrum, and the bottom 10\% floor, where eigenvalues approach numerical noise), applied identically to every model and layer. A more compressed representation decays faster, giving a larger $\alpha$; we report this exponent at the model level (mean across layers) and at the SAE-hook layer specifically. Participation ratio, $(\sum_i\lambda_i)^2/\sum_i\lambda_i^2$, is a second summary of the same eigenvalue spectrum: unlike $\alpha$, it requires no fitting or rank-range choice, so agreement between the two -- both moving the same direction across the sweep -- is evidence the $\alpha$-ReQ trend is not an artifact of the fit range. HT-SR weight-spectrum alpha (WeightWatcher) was also measured and found flat across the full sweep (2.923--2.934) -- the continual-training update does not move the weight spectrum measurably, so this metric is not used further.

\begin{table}[h]
\centering
\caption{Representational and causal metrics across the full qualification sweep. Reconstruction MSE (inference-threshold mode) and IOI candidates are three-seed means (SAE training seeds 42/43/44); $\alpha$-ReQ, participation ratio, and edges-to-85\% are single deterministic measurements, since neither the $\alpha$-ReQ probe (SAE-free, computed directly from the activation covariance) nor circuit recovery (run on the raw graph) depends on SAE training stochasticity. The edges-to-85\% column is the single 85\% gate used for qualification. Representational metrics move in the predicted direction as adversarial strength increases, with occasional non-monotonic steps between adjacent configurations of similar measured robustness (Table~\ref{tab:qualification}); circuit size does not track them at this threshold; the 90\% and 95\% crossings are in Table~\ref{tab:sweep-crossings}.}
\label{tab:dose-response}
\scriptsize
\setlength{\tabcolsep}{3.2pt}
\begin{tabular}{lrrrrrrrr}
\hline
Model & $\epsilon_{\mathrm{rel}}$ & $\alpha$ & Recon.\ MSE & IOI cand.\ & $\alpha$-ReQ (model) & $\alpha$-ReQ (SAE) & Part.\ ratio & Edges$\to$85\% \\
\hline
\texttt{ft-clean}     & 0     & 0   & 1.1528 & 1600.7 & 0.9230 & 0.8074 & 3.815 & 1{,}000 \\
\texttt{e050-a02}     & .050  & .2  & 1.1315 & 1483   & 0.9465 & 0.8177 & 3.689 & \textbf{2{,}000} \\
\texttt{e075-a02}     & .075  & .2  & 1.1277 & 1408   & 0.9546 & 0.8213 & 3.695 & 1{,}000 \\
\texttt{e100-a02}     & .100  & .2  & 1.1242 & 1371   & 0.9609 & 0.8239 & 3.681 & 1{,}000 \\
\texttt{e050-a05}     & .050  & .5  & 1.1432 & 1370   & 0.9644 & 0.8267 & 3.782 & 1{,}000 \\
\texttt{e075-a05}     & .075  & .5  & 1.1301 & 1266   & 0.9762 & 0.8325 & 3.732 & 1{,}000 \\
\texttt{e100-a05}     & .100  & .5  & 1.1260 & 1182   & 0.9865 & 0.8376 & 3.716 & 1{,}000 \\
\hline
\end{tabular}
\end{table}

\section{Corpus ablation: FineWeb}
\label{app:corpus-ablation}

The corpus-ablation pair uses the same continual-training recipe, budget, and qualification gate as the primary standard-robust pair (Section~\ref{sec:continual-training}--\ref{sec:qualification-method}), with FineWeb substituted for OpenWebText as the training and SAE corpus. SAE architecture, training budget, and evaluation protocol match Axis 1 exactly. Both arms pass the IOI qualification floor and the masking sanity check before comparison.

\begin{table}[h]
\centering
\caption{Corpus ablation (inference-threshold mode MSE): standard vs.\ robust GPT-2 Small continually trained on FineWeb in place of OpenWebText, otherwise identical to the primary comparison.}
\label{tab:corpus-ablation}
\small
\begin{tabular}{lrr}
\hline
Metric & Standard (\texttt{fw-clean}) & Robust (\texttt{fw-robust}) \\
\hline
Recon.\ MSE & 1.089 & \textbf{1.077} \\
IOI candidates & 1549 & \textbf{1126} \\
$\alpha$-ReQ (model) & 0.9288 & \textbf{0.9943} \\
$\alpha$-ReQ (SAE) & 0.8121 & \textbf{0.8434} \\
Edges$\to$85\% & 1{,}000 & 1{,}000 \\
Edges$\to$90\% & 4{,}000 & \textbf{1{,}000} \\
Edges$\to$95\% & 8{,}000 & \textbf{2{,}000} \\
\hline
\end{tabular}
\end{table}

\section{Reproducibility checks and limitations}
\label{app:reproducibility}

All experiments use seed 42 for model initialization where applicable, data order, PGD initialization, and primary SAE training; SAE reconstruction and attribution analyses are additionally replicated at seeds 43 and 44. The implementation is covered by 80 CPU tests on small models, including token accounting, attack constraints, value identity of the straight-through splice, attribution-patching linearity against finite differences, BatchTopK selection semantics, KL-control computation, and provenance checks.

The main limitations are the use of one base architecture, one adversarial training regime, and one analyzed residual-stream site. SAE reconstruction and SAE feature engagement use three seeds; the ablation-KL population comparison (Appendix~\ref{app:selectivity-controls}) also uses three seeds but is not seed-robust (one of three IOI seeds reverses direction); circuit recovery (Section~\ref{sec:circuit-results}) uses a 200-prompt subsample, smaller than the 1,000-prompt sets used elsewhere, reflecting its higher per-prompt cost. Circuit recovery is a single deterministic run per model on one fixed prompt subset: we did not run independent prompt resamples or compute bootstrap intervals, and seed-to-seed variation of the standard training run was not measured. The $k$-grid used for circuit size can rule out a large reduction but not distinguish modest differences inside the same bin (Section~\ref{sec:circuit-results}). On Greater-Than the robust model scores lower than the standard model on the same prompts (PD 0.442 vs.\ 0.671; paired difference $-0.229\pm0.010$). We therefore do not treat GT circuit size as a competence-matched test. OpenWebText identifiers are synthesized rather than native. Our EAP-GP implementation follows the method description in the absence of a public reference implementation, with one identified approximation not independently validated against the source paper's reported results.

\end{document}